\def\year{2021}\relax
\documentclass[letterpaper]{article} % DO NOT CHANGE THIS
\usepackage{aaai21}  % DO NOT CHANGE THIS
\usepackage{times}  % DO NOT CHANGE THIS
\usepackage{helvet} % DO NOT CHANGE THIS
\usepackage{courier}  % DO NOT CHANGE THIS
\usepackage[hyphens]{url}  % DO NOT CHANGE THIS
\usepackage{graphicx} % DO NOT CHANGE THIS
\usepackage{natbib}  % DO NOT CHANGE THIS AND DO NOT ADD ANY OPTIONS TO IT
\usepackage{caption} % DO NOT CHANGE THIS AND DO NOT ADD ANY OPTIONS TO IT
\usepackage{algorithm}  
\usepackage{algorithmicx} 
\usepackage{algpseudocode}

\usepackage{multirow}
\usepackage{booktabs}
\usepackage{colortbl} 
\usepackage{arydshln} 

\usepackage{amsmath}
\usepackage{amssymb}
\usepackage{verbatim}
\usepackage{siunitx}
\usepackage{bm}

\title{Multi-View Feature Representation for Dialogue Generation with \\ Bidirectional Distillation}
\author {
        Shaoxiong Feng,\textsuperscript{\rm 1}
        Xuancheng Ren,\textsuperscript{\rm 2}
        Kan Li,\textsuperscript{\rm 1}
        Xu Sun\textsuperscript{\rm 2,3} \\
}
\affiliations {
    \textsuperscript{\rm 1}School of Computer Science \& Technology, Beijing Institute of Technology \\
    \textsuperscript{\rm 2}MOE Key Laboratory of Computational Linguistics, School of EECS, Peking University \\
    \textsuperscript{\rm 3}Center for Data Science, Peking University \\
    \{shaoxiongfeng, likan\}@bit.edu.cn, \{renxc, xusun\}@pku.edu.cn 
}

\begin{document}
\maketitle

\begin{abstract}
Neural dialogue models suffer from low-quality responses when interacted in practice, demonstrating difficulty in generalization beyond training data. Recently, knowledge distillation has been used to successfully regularize the student by transferring knowledge from the teacher. However, the teacher and the student are trained on the same dataset and tend to learn similar feature representations, whereas the most general knowledge should be found through differences. The finding of general knowledge is further hindered by the unidirectional distillation, as the student should obey the teacher and may discard some knowledge that is truly general but refuted by the teacher. To this end, we propose a novel training framework, where the learning of general knowledge is more in line with the idea of reaching consensus, i.e., finding common knowledge that is beneficial to different yet all datasets through diversified learning partners. Concretely, the training task is divided into a group of subtasks with the same number of students. Each student assigned to one subtask not only is optimized on the allocated subtask but also imitates multi-view feature representation aggregated from other students (i.e., student peers), which induces students to capture common knowledge among different subtasks and alleviates the over-fitting of students on the allocated subtasks. To further enhance generalization, we extend the unidirectional distillation to the bidirectional distillation that encourages the student and its student peers to co-evolve by exchanging complementary knowledge with each other. Empirical results and analysis demonstrate that our training framework effectively improves the model generalization without sacrificing training efficiency.
\end{abstract}
\section{Introduction}
Neural dialogue generation has drawn increasing attention, but current dialogue models still struggle with generalization, e.g., frequently producing generic and meaningless responses in inference \cite{DBLP:conf/coling/MouSYL0J16,DBLP:conf/naacl/LiGBGD16,DBLP:conf/aaai/SerbanSLCPCB17}. Unlike machine translation or summarization, dialogue generation has more freedom and diversity in the semantic and linguistic aspects of responses. Without specific training guidance, they are prone to over-fitting certain aspects of corpora (e.g., naive target sequence prediction) that show distinct distributions between training and test data. Therefore, it is usually hard for these models to learn generalizable features and they may easily get stuck in a narrow local minimum that is fragile to data perturbation \cite{DBLP:conf/iclr/ChaudhariCSLBBC17,DBLP:conf/iclr/KeskarMNST17}.

To alleviate this problem, one line of work introduces prior \textit{common knowledge} of the real world to facilitate the model generalization, such as redesigning objective functions (e.g., maximize mutual information or coherence) instead of only fitting target sequence \cite{DBLP:conf/emnlp/LiMRJGG16,DBLP:conf/aaai/FengCLY20}, and modifying generation order (e.g., hierarchical or syntactic-based generation) instead of naive left-to-right generation \cite{DBLP:conf/naacl/SuLYC18,DBLP:conf/icml/WelleckBDC19}. Intuitively, common knowledge is a class of knowledge that benefits both the training and the test data, as it is reflected generally in the whole corpus and not merely only works for the training data. With the common knowledge as constraint, models can be guided to learn towards a better direction that can bridge the gap of training and test data distributions more easily. \citet{DBLP:conf/icml/DubeyAPGE18} also verified that common knowledge from the real world plays an important role in models' quickly learning unfamiliar video games. 

However, prior common knowledge is hard to manually define, since it varies with tasks and domains and can be a limiting factor if defined wrong. Recently, another line of work \cite{DBLP:conf/naacl/AroraKR19,DBLP:conf/sigir/TahamiGS20,DBLP:conf/acl/ChenGCLL20}, using \textit{knowledge distillation} (KD; \citeauthor{DBLP:conf/nips/BaC14}~\citeyear{DBLP:conf/nips/BaC14}; \citeauthor{DBLP:journals/corr/HintonVD15}~\citeyear{DBLP:journals/corr/HintonVD15}), has successfully extracted knowledge from a pre-trained teacher model to regularize the student model for better generalization. The student model aims to achieve a balance of using raw knowledge from the training data and distilled knowledge from the teacher model, which makes the student capture more common or generalizable knowledge and perform better in testing. Compared with previous work on introducing common knowledge, KD is more straightforward and extensible. However, conventional KD still faces two drawbacks:
\begin{itemize}
    \item Lack of feature diversity: Because both the student and the teacher are trained on the same dataset, they may learn similar feature representations, which means the knowledge is not sufficiently diverse to conduct an effective regularization on the feature learning of the student.
    \item Lack of student feedback: Previous work \cite{DBLP:journals/corr/RomeroBKCGB14,DBLP:conf/cvpr/YimJBK17,DBLP:conf/icml/FurlanelloLTIA18} has proved that the student can obtain better generalization performance than the teacher, but KD still runs unidirectionally, which may damage the more generalizable knowledge learned by the student and hinder the performance improvement of the teacher.
\end{itemize}

\begin{figure}[t]
    \centering
    \includegraphics[width=1.0\linewidth]{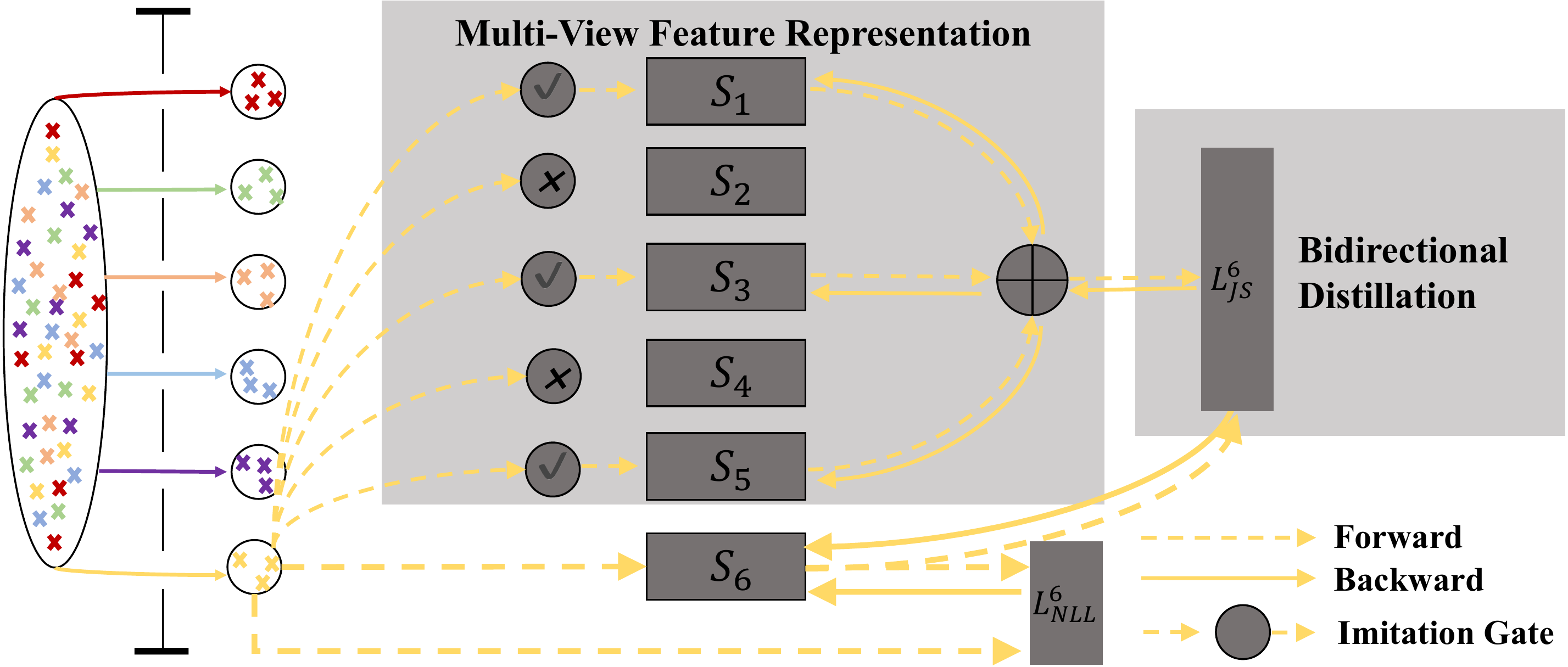}
    \caption{\textit{An overview of multi-view feature representation and bidirectional distillation.}}
    \label{fig:architecture}
    % \vspace{-1.0\baselineskip}
\end{figure}

In this work, we propose a novel training framework to tackle these problems by generating multi-view feature representations and co-evolution via bidirectional distillation. Figure~\ref{fig:architecture} illustrates the proposed framework. To obtain multi-view feature representations, the training task is divided into a group of subtasks, i.e., subsets of training data, and each subtask is assigned a corresponding student model, which learns knowledge specific to different subtasks. The students are also enforced to perform as well as possible in the unseen subtasks by imitating the predictions of other students, so that common knowledge can be drawn collaboratively, which also prevents aggressive over-fitting. In addition, a bidirectional knowledge distillation is further applied, which encourages the student and its student peers to exchange complementary knowledge and together evolve towards better generalization, which further eliminates the need of pre-training in conventional KD. Furthermore, the student peers for a student in knowledge distillation are randomly selected at each iteration to prevent the degeneration and homogenization \cite{DBLP:journals/ml/KunchevaW03,DBLP:journals/cim/Schwenker13} of multi-view feature representations due to the bidirectional learning settings. In such way, the proposed training framework enables students to jointly learn diverse yet generalizable knowledge from multi-view feature representations from different data compositions.

Our main contributions are as follows:
\begin{itemize}
\item We propose a novel training framework that reconstructs the training task as a group of subtasks and aggregates multi-view feature representations from randomly-selected student peers to regularize students for more generalizable knowledge.
\item The framework is enhanced by bidirectional knowledge distillation that allows the student to provide feedback to its student peers and makes both ends able to co-evolve.
\item We conducted extensive experiments and analysis to validate the effectiveness of multi-view feature representations and bidirectional distillation and demonstrate why these mechanisms work well.
\end{itemize}
\section{Method}
In this section, we describe how to effectively capture common knowledge for improving the generalization of dialogue models. We first introduce multi-view feature representation for diverse knowledge, then propose bidirectional distillation to regularize both students and their partners, and finally present the optimization objective.

%%%%%%%%%%%%%%%%%%%%%%%%%%%%%%%%%%%%%%%%%%%%%%%%%%%%%%%%%%%%%%%%%%%%%%%%%%%%%%%%%%
%%%%%%%%%%%%%%%%%%%%%%%%%%%%%%%%%%%%%%%%%%%%%%%%%%%%%%%%%%%%%%%%%%%%%%%%%%%%%%%%%%

\subsection{Multi-View Feature Representation}
For generative conversation models, given a training example $(x, y) \in \mathcal{D}$, where $x$ is the source sequence, i.e., the dialogue history, $y$ is the target sequence, i.e., the corresponding response, and $\mathcal{D}$ is the whole training dataset consisting of $M$ examples, i.e., $|\mathcal{D}|=M$, the learning objective is to minimize the following negative log-likelihood:
%%%%%%%%%%%%%%%%%%%%%%%%%%%%%%%%%%%%%%%%%%%%%%%%%%%%%%%%%%%%%%%%%%%%%%%%%%%
\begin{equation}
\small
\begin{aligned}
\mathcal{L}_{\mathrm{NLL}}(x, y; \theta) = - \sum\nolimits_{j=1}^{|y|} \log p(y_{j} |y_{<j}, x; \theta), 
\end{aligned}
\label{equation: traditional student NLL}
\end{equation}
%%%%%%%%%%%%%%%%%%%%%%%%%%%%%%%%%%%%%%%%%%%%%%%%%%%%%%%%%%%%%%%%%%%%%%%%%%%
where $\theta$ is the learnable parameters.

In conventional knowledge distillation \cite{DBLP:conf/icml/FurlanelloLTIA18,DBLP:journals/corr/abs-2006-12000}, a teacher model is first pre-trained on the whole dataset, whose parameters we denote as $\theta_t$, and another model, parameterized as $\theta_s$, is taken as the student that further aligns its prediction with the teacher prediction using the Kullback-Leibler divergence \cite{kullback1951information}:
%%%%%%%%%%%%%%%%%%%%%%%%%%%%%%%%%%%%%%%%%%%%%%%%%%%%%%%%%%%%%%%%%%%%%%%%%%%
% \begin{equation}
$$
\small 
\begin{aligned}
\mathcal{L}_{\mathrm{KL}}(x, y, \theta_{t}; \theta_{s}) 
& = \sum_{j=1}^{|y|} \sum_{w \in \mathcal{V}} p(w|X_j; \theta_t) \log \frac{p(w|X_j; \theta_t)}{p(w|X_j; \theta_s)},
\end{aligned}
\label{equation: traditional KD}
$$
% \end{equation}
%%%%%%%%%%%%%%%%%%%%%%%%%%%%%%%%%%%%%%%%%%%%%%%%%%%%%%%%%%%%%%%%%%%%%%%%%%%
where $w$ is a word in the vocabulary $\mathcal{V}$ and $X_j$ is defined as $(y_{<j}, x)$. The student is tasked to cover all the knowledge from the teacher due to the mean-seeking behaviour of the KL divergence, while the teacher is kept fixed in the distillation.

As the teacher and the student are trained using exactly the same data, they intend to learn similar feature representations or knowledge \cite{DBLP:journals/corr/LiYCLH15,DBLP:conf/nips/MorcosRB18}. However, it should be crucial for the students to obtain sufficiently diverse sources of knowledge to extract common knowledge that generalizes to unseen examples, which is not available in such distillation settings and limits the effect of the regularization from the teacher. To address this problem, we propose to learn multi-view feature representations for the students to find more common knowledge by aligning their predictions with diverse partners. Essentially, the training dataset is broken down into $N$ subsets $\{\mathcal{D}^n\}_{n=1}^{N}$, where $\cup_{n=1}^{N} \mathcal{D}^n = \mathcal{D}$ and $\cap_{n=1}^{N} \mathcal{D}^n = \varnothing$, that compose varied subtasks, each assigned an individual student. Each student is trained using supervised examples solely from its corresponding subset so that for the macro task we can get diverse representations from micros views. The supervised learning of a student is conducted as follows:
%%%%%%%%%%%%%%%%%%%%%%%%%%%%%%%%%%%%%%%%%%%%%%%%%%%%%%%%%%%%%%%%%%%%%%%%%%%
\begin{equation}
\small
\begin{aligned}
\mathcal{L}^{n}_{\mathrm{NLL}}(x^{k}, y^{k}; \theta_{n}) = - \sum_{j=1}^{|y^{k}|} \log p(y^{k}_j|y^{k}_{<j}, x^{k}; \theta_{n}),
\end{aligned}
\label{equation: new student NLL}
\end{equation}
%%%%%%%%%%%%%%%%%%%%%%%%%%%%%%%%%%%%%%%%%%%%%%%%%%%%%%%%%%%%%%%%%%%%%%%%%%%
where $(x^{k}, y^{k})\in \mathcal{D}^n$ and $n$ identifies the student and the subset.

In turn, we aggregate the corresponding multi-view feature representation for each student to imitate by averaging over predictions of all other students. However, aggregation with naive averaging will lead multi-view feature representations of all students to be similar, which can also cause students to homogenize \cite{DBLP:journals/ml/KunchevaW03,DBLP:journals/cim/Schwenker13}. Therefore, we further introduce the imitation gate $g(\cdot)$ (shown in Figure \ref{fig:architecture}) to maintain the diversity of multi-view feature representations. Specifically, each student randomly imitates a subgroup of students at each iteration, which subjects to $g(p) \sim \text{Bernoulli}(p)$, and the number of the imitated students is decided by the imitation probability $p$. During training, students are regularized by different and dynamic multi-view feature representations, which alleviates the homogenization of students. Besides, the imitation gate also reduces the computational cost (i.e., forward and backward propagation). For example, we can keep the computational cost constant by adjusting the imitation probability when the number of students increases. 
The distillation loss of each student is computed as:
%%%%%%%%%%%%%%%%%%%%%%%%%%%%%%%%%%%%%%%%%%%%%%%%%%%%%%%%%%%%%%%%%%%%%%%%%%%
% \begin{equation}
$$
\small 
\begin{aligned}
\mathcal{L}^{n}_{\mathrm{KL}}(x^{k}, y^k, \theta_{/n}; \theta_{n}) =  
& \sum_{j=1}^{|y^k|} \sum_{w \in \mathcal{V}} p(w|X^k_j; \theta_{/n}) \frac{p(w|X^k_j; \theta_{/n})}{p(w|X^k_j; \theta_n)},
\end{aligned}
\label{equation: new student KD}
$$
% \end{equation}
%%%%%%%%%%%%%%%%%%%%%%%%%%%%%%%%%%%%%%%%%%%%%%%%%%%%%%%%%%%%%%%%%%%%%%%%%%% 
where $p(w|X^k_j; \theta_{/n})$ is the aggregated probability distribution from other students, $/n$ denotes the students other than the student $n$, and $X_j^k$ denotes $(y^k_{<j}, x^k) \in \mathcal{D}^n$, which ensures that the student $n$ never observes data directly from other subsets. $p(w|X^k_j; \theta_{/n})$ is calculated as:
%%%%%%%%%%%%%%%%%%%%%%%%%%%%%%%%%%%%%%%%%%%%%%%%%%%%%%%%%%%%%%%%%%%%%%%%%%%
\begin{equation}
\small 
p(w | X_j^k; \theta_{/n}) \triangleq \frac{1}{H} \sum_{i=1 \ldots N, i \neq n} g^i(p) 
p(w|X_j^k; \theta_i),
\label{equation: imitation gate}
\end{equation}
%%%%%%%%%%%%%%%%%%%%%%%%%%%%%%%%%%%%%%%%%%%%%%%%%%%%%%%%%%%%%%%%%%%%%%%%%%%
where $H = \sum_{i=1...N, i \neq n} g^i(p)$ is the number of the imitated students. By randomly selecting distilled students at each iteration, a single student is kept from the access of a global view of all the data and may maintain its specialty rather than homogenize.

%%%%%%%%%%%%%%%%%%%%%%%%%%%%%%%%%%%%%%%%%%%%%%%%%%%%%%%%%%%%%%%%%%%%%%%%%%%%%%%%%%
\begin{figure}[t]
    \centering
    \includegraphics[width=1.0\linewidth]{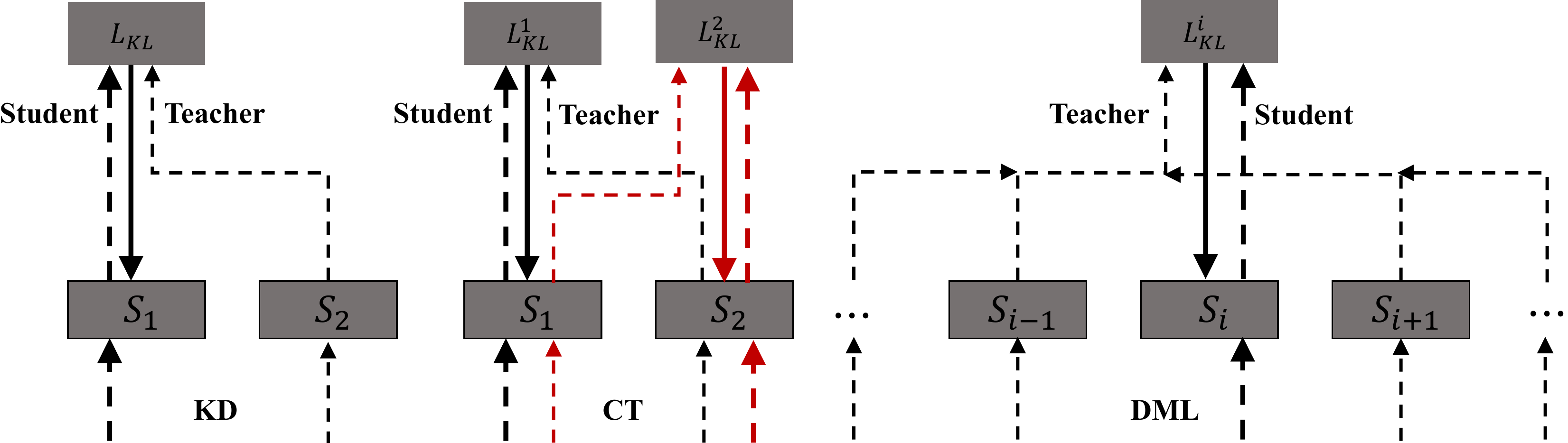}
    \caption{Comparison of knowledge aggregation and transfer among vanilla KD, Co-Teaching (CT), and deep mutual learning (DML). The dashed and solid lines represent forward and backward propagation, respectively. The differently colored lines in CT represent students $S_1$ and $S_{2}$ are trained on independent batches of training data.}
    \label{fig:KD-CT-DML}
    % \vspace{-1.0\baselineskip}
\end{figure}
%%%%%%%%%%%%%%%%%%%%%%%%%%%%%%%%%%%%%%%%%%%%%%%%%%%%%%%%%%%%%%%%%%%%%%%%%%%%%%%%%%

\subsection{Bidirectional Distillation}
Previous work, such as vanilla KD \cite{DBLP:journals/corr/HintonVD15}, Co-Teaching (CT) \cite{DBLP:conf/acl/FengTWFZY19}, and deep mutual learning (DML) \cite{DBLP:conf/cvpr/ZhangXHL18}, only conducts the unidirectional knowledge distillation from the teacher to the student, illustrated in Figure \ref{fig:KD-CT-DML}. However, according to prior research \cite{DBLP:journals/corr/RomeroBKCGB14,DBLP:conf/cvpr/YimJBK17,DBLP:conf/icml/FurlanelloLTIA18}, the student can achieve better performance than the teacher. It suggests the unidirectional knowledge distillation has the risk of damaging the more generalizable knowledge that the student has learned. 

To alleviate this problem, we propose the bidirectional knowledge distillation that induces both sides to reach a consensus by simultaneously regularizing each other, rather than one side monotonously imitating the other side. Similar to feature fusion methods \cite{DBLP:conf/iccv/HouLW17}, we first fuse the prediction $p(w|X^k_j; \theta_n)$ of the student $n$ and the prediction $p(w|X^k_j; \theta_{/n})$ of corresponding multi-view feature representation to construct more generalizable knowledge $p(w|X^k_j; \theta_n,\theta_{/n})$:
%%%%%%%%%%%%%%%%%%%%%%%%%%%%%%%%%%%%%%%%%%%%%%%%%%%%%%%%%%%%%%%%%%%%%%%%%%%
\begin{equation}
\small 
p(w|X^k_j; \theta_n,\theta_{/n}) \triangleq 
\frac{1}{2} p(w|X^k_j; \theta_n) + 
\frac{1}{2} p(w|X^k_j; \theta_{/n})
\label{equation: temp term for JS}
\end{equation}
%%%%%%%%%%%%%%%%%%%%%%%%%%%%%%%%%%%%%%%%%%%%%%%%%%%%%%%%%%%%%%%%%%%%%%%%%%%
Then we enforce the student to imitate the fused knowledge $p(w|X^k_j; \theta_n,\theta_{/n})$, which can be expressed as:
%%%%%%%%%%%%%%%%%%%%%%%%%%%%%%%%%%%%%%%%%%%%%%%%%%%%%%%%%%%%%%%%%%%%%%%%%%%
\begin{equation}
\small 
\begin{aligned}
\mathcal{L}^{n}_{\mathrm{KL}}(x^{k}, y^k, \theta_{/n}; \theta_{n}) =  
& \sum_{j=1}^{|y^k|} \sum_{w \in \mathcal{V}}
p(w|X^k_j; \theta_n,\theta_{/n}) \cdot \\
& \log \frac{p(w|X^k_j; \theta_n,\theta_{/n})}{p(w|X^k_j; \theta_n)} 
\end{aligned}
\label{equation: new student KD for JS}
\end{equation}
%%%%%%%%%%%%%%%%%%%%%%%%%%%%%%%%%%%%%%%%%%%%%%%%%%%%%%%%%%%%%%%%%%%%%%%%%%%
Besides, we also allow the teacher (i.e., the partners) to be regularized and conduct parameter updates. The final distillation loss is formulated as the Jensen–Shannon (JS) divergence \cite{DBLP:conf/acl/DaganLP97}: 
%%%%%%%%%%%%%%%%%%%%%%%%%%%%%%%%%%%%%%%%%%%%%%%%%%%%%%%%%%%%%%%%%%%%%%%%%%%
\begin{equation}
\small 
\begin{aligned}
\mathcal{L}^{n}_{\mathrm{JS}}(x^{k}, y^k; \theta_{n}, \theta_{/n}) = 
& \frac{1}{2} \mathcal{L}^{n}_{\mathrm{KL}}(x^{k}, y^k, \theta_{/n}; \theta_{n}) + \\
& \frac{1}{2} \mathcal{L}^{n}_{\mathrm{KL}}(x^{k}, y^k, \theta_{n}; \theta_{/n})
\end{aligned}
\label{equation: student JS}
\end{equation}
%%%%%%%%%%%%%%%%%%%%%%%%%%%%%%%%%%%%%%%%%%%%%%%%%%%%%%%%%%%%%%%%%%%%%%%%%%% 
In CT and DML, students can also provide knowledge for the imitated student due to iterative parameter updates. However, iterative parameter updates have two disadvantages: 1) It obviously slows down the training speed. 2) Compared with it, simultaneous parameter updates can speed up the convergence and obtain better performance \cite{DBLP:conf/nips/MeschederNG17,DBLP:conf/nips/NagarajanK17}.

\subsection{Optimization}
In this section, combining the NLL loss in Equation \ref{equation: new student NLL} with the distillation loss in Equation \ref{equation: student JS}, we give the final optimization objective as follows: 
\begin{equation}
\small
\mathcal{L} = \sum_{n=1}^{N} ( \mathcal{L}^{n}_{\mathrm{NLL}} + T^{2}*\mathcal{L}^{n}_{\mathrm{JS}} ),
\end{equation}
where $T^{2}$ is a scalar coefficient. It is used to maintain an equilibrium between the NLL loss $\mathcal{L}^{n}_{\mathrm{NLL}}$ and the distillation loss $\mathcal{L}^{n}_{\mathrm{JS}}$ because we use $\frac{1}{T}$ to soften the probability distribution when calculating KL divergence. 

Unlike vanilla KD and CT, we do not pre-train any student model as DML, which significantly saves the training time. Once the whole training set is divided into a group of subtasks, each student learns the assigned subtask and conducts the bidirectional distillation simultaneously. All students update parameters in parallel until convergence. 
\section{Experiment}
We conduct experiments on two high-quality open-domain dialogue datasets, DailyDialog and PersonaChat, compared with state-of-the-art methods, and provide extensive analysis to examine the effect of the proposed method.

\subsection{Datasets}
We adopt two commonly-used dialogue datasets:
\begin{itemize}
\item \textbf{DailyDialog} \cite{DBLP:conf/ijcnlp/LiSSLCN17} covers a variety of daily scenarios, such as work, health, and politics. 
We first extract the \textit{(dialogue history, response)} pairs from the raw dataset. Each pair consists of two consecutive dialogue turns, in which the first turn and the second turn represent dialogue history and response, respectively. 
Then we limit the length of dialogue turns to $[5, 25]$ by discarding the pairs whose response is shorter than 5 words and truncating the turns whose length is longer than 25 words.
Finally, the processed dataset contains 50K, 4.5K, and 4.3K pairs for training, validation, and testing, respectively. 

\item \textbf{PersonaChat} \cite{DBLP:conf/acl/KielaWZDUS18} is collected by two crowdsourced workers chit-chatting with each other, conditioned on the assigned personas. 
In our experiments, we only use the conversation text and process it as DailyDialog.
The processed dataset contains 106K, 13K, and 12.5K pairs for training, validation, and testing.
\end{itemize}

\subsection{Baselines}
We re-implemented the following four methods and compared them with the proposed \textbf{MRBD} (\textbf{M}ulti-View Feature \textbf{R}epresentation and \textbf{B}idirectional \textbf{D}istillation): 
\begin{itemize}
\item \textbf{Seq2Seq+Att} uses a vanilla Seq2Seq model \cite{DBLP:conf/nips/SutskeverVL14} with attention mechanism \cite{DBLP:journals/corr/BahdanauCB14}. The encoder and the decoder are based on a 2-layer bidirectional GRU \cite{DBLP:conf/emnlp/ChoMGBBSB14} and a 2-layer unidirectional GRU, respectively. The size of hidden units is 500.

\item \textbf{KD} uses two dialogue models as the student and the teacher, similar to \citet{DBLP:conf/sigir/TahamiGS20}. The student learns from both the ground-truth responses and the probability distributions of the teacher.

\item \textbf{CT} stands for the Co-Teaching training framework \cite{DBLP:conf/acl/FengTWFZY19}, in which two students are trained on independent training sets, and they provide complementary knowledge for each other. In practice, one student can still access the data assigned to the other student due to the whole training set shuffled once in each epoch. 

\item \textbf{DML} means Deep Mutual Learning \cite{DBLP:conf/cvpr/ZhangXHL18}, which constructs a group of students trained on the same training set. Each student learns from both the ground-truth responses and the knowledge equally aggregated from all other students. All students update the parameters iteratively, which means one student needs to recalculate a new prediction for the next students to imitate after updating its parameters. 
\end{itemize} 
In our experiments, for a fair comparison, models in baselines and our method comprise the Seq2Seq-based generative dialogue models with the same settings as Seq2Seq+Att. Besides, we use the KL divergence to calculate the distance of probability distributions for all baselines.

\subsection{Experimental settings}
According to the performance on the validation set, including loss and metrics, we set the hyper-parameters of the proposed method and baselines as follows: 
We set the embedding size to 500, the vocabulary size for both DailyDialog and PersonaChat to 20K. 
The dropout probability and the temperature $T$ are 0.1 and 3, respectively. 
We use Adam optimizer \cite{DBLP:journals/corr/KingmaB14}, with a learning rate of 0.0001, gradient clipping at 5.0, and a mini-batch size of 64. 
Following the settings of \citet{DBLP:conf/acl/FengTWFZY19}, CT needs to pre-train students on the whole training set before Co-Teaching. 
We set the number of students to 6 for DML and MRBD. 
The imitation probability in MRBD is 0.5. 
The training set is randomly divided into six non-overlapping subsets with the same number of pairs.
For CT, DML, and MRBD, we choose the student model that achieves the best performance on the validation set for the final evaluation.

%%%%%%%%%%%%%%%%%%%%%%%%%%%%%%%%%%%%%%%%%%%%%%%%%%%%%%%%%%%%%%%%%%%%%%%%%%%%
\begin{table}[t]
\centering
\footnotesize
\setlength{\tabcolsep}{4.0pt}

\begin{tabular}{@{}lcccccc@{}}
\hline
%\multicolumn{7}{c}{DailyDialog} \\ \hline
DailyDialog & Dist-1 & Dist-2 & Ent-1 & Ent-2 & Dis-1 & Dis-2 \\ \hline
Seq2Seq+Att & 4.054 & 27.962 & 7.689 & 12.773 & 0.148 & 0.454 \\
KD & 4.219 & 29.007 & 7.732 & 12.892 & 0.142 & 0.395 \\
CT & 4.591 & 29.387 & 7.864 & 13.087 & 0.323 & 0.477 \\
DML & 4.316 & 29.193 & 8.027 & 12.932 & 0.146 & 0.374 \\
MRBD & \bf 4.762 & \bf 30.592 & \bf 8.232 & \bf 13.257 & \bf 0.136 & \bf 0.357 \\ \hline\hline
%\multicolumn{7}{c}{PersonaChat} \\ \hline
PersonaChat & Dist-1 & Dist-2 & Ent-1 & Ent-2 & Dis-1 & Dis-2 \\ \hline
Seq2Seq+Att & 0.854 & 5.122 & 7.136 & 11.294 & 0.601 & 1.138 \\
KD & 0.862 & 5.343 & 7.159 & 11.718 & 0.502 & 0.964 \\
CT & 1.093 & 7.161 & 7.233 & 12.038 & 0.641 & 1.246 \\
DML & 0.952 & 6.399 & 7.196 & 11.824 & 0.435 & 0.891 \\
MRBD & \bf 1.745 & \bf 12.391 & \bf 7.419 & \bf 12.246 & \bf 0.300 & \bf 0.578 \\ \hline
\end{tabular}
\caption{Results of the automatic evaluation.}
\label{experiment result: automatic evaluation}
% \vspace{-0.5\baselineskip}
\end{table}
%%%%%%%%%%%%%%%%%%%%%%%%%%%%%%%%%%%%%%%%%%%%%%%%%%%%%%%%%%%%%%%%%%%%%%%%%%%%
\subsection{Experimental Results}

It is challenging to assess the quality of the generated responses, especially in semantics (e.g., coherence and fluency).
In this work, we conduct two kinds of evaluations, automatic evaluation and human evaluation. 
The automatic evaluation focuses on the diversity, specificity, and distribution of responses that can be well reflected by the statistics of words. 
The human evaluation considers the coherence, similarity, and fluency of responses. Both BLEU \cite{DBLP:conf/acl/PapineniRWZ02} and EmbSim \cite{DBLP:conf/emnlp/LiuLSNCP16} are adopted to measure the similarity of the generated response with reference, but they show a poor correlation with human evaluation. 

\subsubsection{Automatic Evaluation} 
\textbf{Dist-\{1,2\}} (Distinct) are widely employed to evaluate the diversity of the generated responses \cite{DBLP:conf/naacl/LiGBGD16,DBLP:conf/nips/ZhangGGGLBD18,DBLP:conf/aaai/FengCLY20}, which represent the percentage (\%) of unique unigrams and bigrams. 
We use \textbf{Ent-\{1,2\}}\footnote{$\text{Ent}=-\frac{1}{|U|} \sum_{w \in U} \log _{2} p_{g}(w)$, where $p_{g}$ is estimated based on the training set.} (Word Entropy) and \textbf{Dis-\{1,2\}}\footnote{$\text{Dis}=\frac{1}{\left|U_{r}\right|} \sum_{w \in U_{r}} \log_{2} \frac{p_{r}(w)}{p(w)}$, where $p_{r}$ and $p$ are estimated based on references and the generated responses, respectively.} (KL divergence) to measure the specificity and distribution distance of the generated responses \cite{DBLP:conf/acl/CsakyPR19}. The responses with higher word entropy contain more meaningful and low-frequency words. A lower KL divergence represents a more similar response distribution. We report both unigrams and bigrams versions of word entropy and KL divergence.
The results are shown in Table \ref{experiment result: automatic evaluation}. 
We can see that our training framework significantly outperforms all state-of-the-art baselines in terms of diversity, specificity, and distribution distance on all datasets, especially on PersonaChat. 
Compared with other baselines, CT also obtains more diverse and more specific responses as MRBD but shows a dramatic decline in the distribution distance of responses. We argue that the diversified multi-view knowledge has better regularization effects than the diversified single-view knowledge for fitting the distribution of the real-world responses. In practice, the performance of CT will weaken once we do not pre-train the students before Co-Teaching. 
Moreover, DML gains more performance improvements than KD in comparison to Seq2Seq+Att, which also validates multi-view knowledge is beneficial for regularizing the feature learning of students.
Finally, we conducted the significant test on both DailyDialog and PersonaChat, and the results demonstrate that the performance improvements of MRBD are significant (i.e., $p<0.01$).

%%%%%%%%%%%%%%%%%%%%%%%%%%%%%%%%%%%%%%%%%%%%%%%%%%%%%%%%%%%%%%%%%%%%%%%%%%%%
\begin{table}[t]
\centering
\footnotesize
\setlength{\tabcolsep}{5.5pt}

\begin{tabular}{@{}lcccc@{}}
\hline
%\multicolumn{5}{c}{DailyDialog} \\ \hline
DailyDialog & Coherence & Similarity & Fluency & Average \\ \hline
Seq2Seq+Att & 3.36 & 2.91 & 3.64 & 3.303 \\
KD & 2.55 & 2.09 & 2.45 & 2.363 \\
CT & 2.45 & 2.18 & 1.82 & 2.150 \\
DML & 2.00 & 1.72 & 2.00 & 1.906 \\
MRBD & \bf 1.45 & \bf 1.55 & \bf 1.09 & \bf 1.363 \\ \hline \hline
%\multicolumn{5}{c}{PersonaChat} \\ \hline
PersonaChat & Coherence & Similarity & Fluency & Average \\ \hline
Seq2Seq+Att & 3.09 & 2.63 & 3.26 & 2.993 \\
KD & 2.25 & 2.13 & 3.01 & 2.463 \\
CT & 2.38 & 1.87 & 2.38 & 2.210 \\
DML & 1.37 & 1.75 & 2.12 & 1.746 \\
MRBD & \bf 1.12 & \bf 1.38 & \bf 1.25 & \bf 1.250 \\ \hline
\end{tabular}
\caption{Results of the human evaluation. Lower is better.}
\label{experiment result: human evaluation}
% \vspace{-0.5\baselineskip}
\end{table}
%%%%%%%%%%%%%%%%%%%%%%%%%%%%%%%%%%%%%%%%%%%%%%%%%%%%%%%%%%%%%%%%%%%%%%%%%%%%
\subsubsection{Human Evaluation} 
For all datasets, we randomly extracted 200 pairs from the test sets. Then we invited three well-educated annotators to rank the responses generated by different models in terms of \textbf{coherence} (how much information in the generated response is relevant to dialogue history), \textbf{similarity} (how much information in the generated response is related to reference), and \textbf{fluency} (how likely the generated response is from human). Ties are allowed. 
Table \ref{experiment result: human evaluation} reports the evaluation results. 
We can see that MRBD achieves consistent improvements across all metrics.
Especially in the coherence and fluency, MRBD shows substantive gains. 
CT and DML have greater advantages than KD with respect to fluency. 
We also calculate the spearman’s rank correlation coefficient \cite{zar2014spearman} to evaluate the inter-annotator agreement. The results are 0.542 and 0.602 on DailyDialog and PersonaChat, respectively, with $p<0.001$.

\subsection{Experimental analysis}
In this section, we provide extensive analysis to validate the effectiveness of multi-view feature representation and bidirectional distillation, and further discuss why the proposed framework works better. Unless otherwise stated, the following results are based on the test set of DailyDialog. 

%%%%%%%%%%%%%%%%%%%%%%%%%%%%%%%%%%%%%%%%%%%%%%%%%%%%%%%%%%%%%%%%%%%%%%%%%%%%
\begin{table}[t]
\centering
\footnotesize
\setlength{\tabcolsep}{4.0pt}

\begin{tabular}{@{}lcccccc@{}}
\hline
Model & Dist-1 & Dist-2 & Ent-1 & Ent-2 & Dis-1 & Dis-2 \\ \hline
w/o Subtask & 5.317 & 32.149 & 7.265 & 12.476 & 0.205 & 0.486 \\
w/o Subgroup & 4.692 & 30.385 & 8.101 & 12.889 & 0.138 & 0.404 \\
w/o BiDistill & 3.836 & 25.480 & 8.225 & 13.449 & 0.134 & 0.348 \\ \hline
\end{tabular}
\caption{Results of the ablation study.}
\label{Result Analysis: Ablation Study}
% \vspace{-0.5\baselineskip}
\end{table}
%%%%%%%%%%%%%%%%%%%%%%%%%%%%%%%%%%%%%%%%%%%%%%%%%%%%%%%%%%%%%%%%%%%%%%%%%%%%
\subsubsection{Ablation Study} 
We first conduct the ablation study to analyze the contributions of different mechanisms quantitatively. Then we further investigate the impact of the overlapping ratio of subtasks and the imitation probability of students on model performance. 

Table \ref{Result Analysis: Ablation Study} shows the results of MRBD w/o Subtask (i.e., students are trained on the same training set), w/o Subgroup (each student imitates all other students), w/o BiDistill (i.e., students adopt unidirectional distillation). 
As we can see, MRBD w/o Subtask improves the diversity of responses but yields a sharp decline in terms of specificity and distribution distance. 
It is because students can not provide diversified multi-view knowledge to regularize each other for common knowledge without the subtask mechanism. 
The generated responses are more diverse but limited in the training set, which is in line with observations in \citet{DBLP:conf/acl/CsakyPR19}, i.e., the diversity of responses still increases after over-fitting the training set. 
MRBD w/o Subgroup shows a slight decrease in all metrics compared with MRBD, which indicates that the subgroup mechanism conducts a positive effect on maintaining the diversity of knowledge. 
The specificity and distribution distance of MRBD w/o BiDistill obtain slight improvements, but the diversity declines dramatically. 
This phenomenon demonstrates that the unidirectional distillation causes students only to imitate the aggregated knowledge and may lose knowledge learned from the assigned subtask.

%%%%%%%%%%%%%%%%%%%%%%%%%%%%%%%%%%%%%%%%%%%%%%%%%%%%%%%%%%%%%%%%%%%%%%%%%%%%
\begin{table}[t]
\centering
\footnotesize
\setlength{\tabcolsep}{4.0pt}

\begin{tabular}{@{}lcccccc@{}}
\hline
Ratio & Dist-1 & Dist-2 & Ent-1 & Ent-2 & Dis-1 & Dis-2 \\ \hline
0\% & 4.762 & 30.592 & 8.232 & 13.257 & 0.136 & 0.357 \\
25\% & 4.808 & 31.585 & 8.351 & 13.343 & 0.133 & 0.308 \\
50\% & 5.115 & 31.191 & 7.311 & 12.550 & 0.167 & 0.498 \\
100\% & 5.317 & 32.149 & 7.265 & 12.476 & 0.205 & 0.486 \\ \hline
\end{tabular}
\caption{Results of different ratios of subtask overlap.}
\label{Result Analysis: Subtask Overlap}
% \vspace{-0.5\baselineskip}
\end{table}
%%%%%%%%%%%%%%%%%%%%%%%%%%%%%%%%%%%%%%%%%%%%%%%%%%%%%%%%%%%%%%%%%%%%%%%%%%%%
\textbf{Impact of Subtask Overlap}  Table \ref{Result Analysis: Subtask Overlap} gives the results of MRBD with the overlapping ratios of 0\%, 25\%, 50\%, 100\%. We can discover that MRBD (25\%) achieves better performance than other variants. 
After the overlapping ratio of 25\%, the performance of MRBD represents a gradual decline in specificity and distribution distance, which is consistent with the observation in MRBD w/o subtask. 
It suggests that allowing subtasks to overlap appropriately is beneficial for students to gain more performance improvements.

%%%%%%%%%%%%%%%%%%%%%%%%%%%%%%%%%%%%%%%%%%%%%%%%%%%%%%%%%%%%%%%%%%%%%%%%%%%%
\begin{table}[t]
\centering
\footnotesize
\setlength{\tabcolsep}{4.0pt}

\begin{tabular}{@{}lcccccc@{}}
\hline
Probability & Dist-1 & Dist-2 & Ent-1 & Ent-2 & Dis-1 & Dis-2 \\ \hline
0.2 & 4.610 & 29.913 & 8.032 & 12.926 & 0.162 & 0.400 \\
0.5 & 4.762 & 30.592 & 8.232 & 13.257 & 0.136 & 0.357 \\
0.8 & 4.841 & 31.219 & 8.138 & 13.013 & 0.136 & 0.364 \\
1.0 & 4.692 & 30.385 & 8.101 & 12.889 & 0.138 & 0.404 \\ \hline
\end{tabular}
\caption{Results of different imitation probabilities.}
\label{Result Analysis: Imitation Probability}
% \vspace{-0.5\baselineskip}
\end{table}
%%%%%%%%%%%%%%%%%%%%%%%%%%%%%%%%%%%%%%%%%%%%%%%%%%%%%%%%%%%%%%%%%%%%%%%%%%%%
\textbf{Impact of Imitation Probability} Table \ref{Result Analysis: Imitation Probability} presents the results of MRBD with the imitation probabilities of 0.2, 0.5, 0.8, and 1.0. The performance of MRBD first ascends and then slowly declines as the imitation probability gradually increases, which means that if students share too many views, the aggregated multi-view knowledge will be more similar, exacerbating the homogenization of students. 
Besides, MRBD consumes less computational cost compared with DML due to the adjustable imitation probability.

\subsubsection{Model Generalization Analysis}
We first validate the robustness of MRBD against noisy data, and then investigate why it achieves better generalization than baselines.

%%%%%%%%%%%%%%%%%%%%%%%%%%%%%%%%%%%%%%%%%%%%%%%%%%%%%%%%%%%%%%%%%%%%%%%%%%%%%%%%%%%%%%%%%%%%%%%
\begin{figure}[t]
    \centering 
    \includegraphics[width=0.53\linewidth]{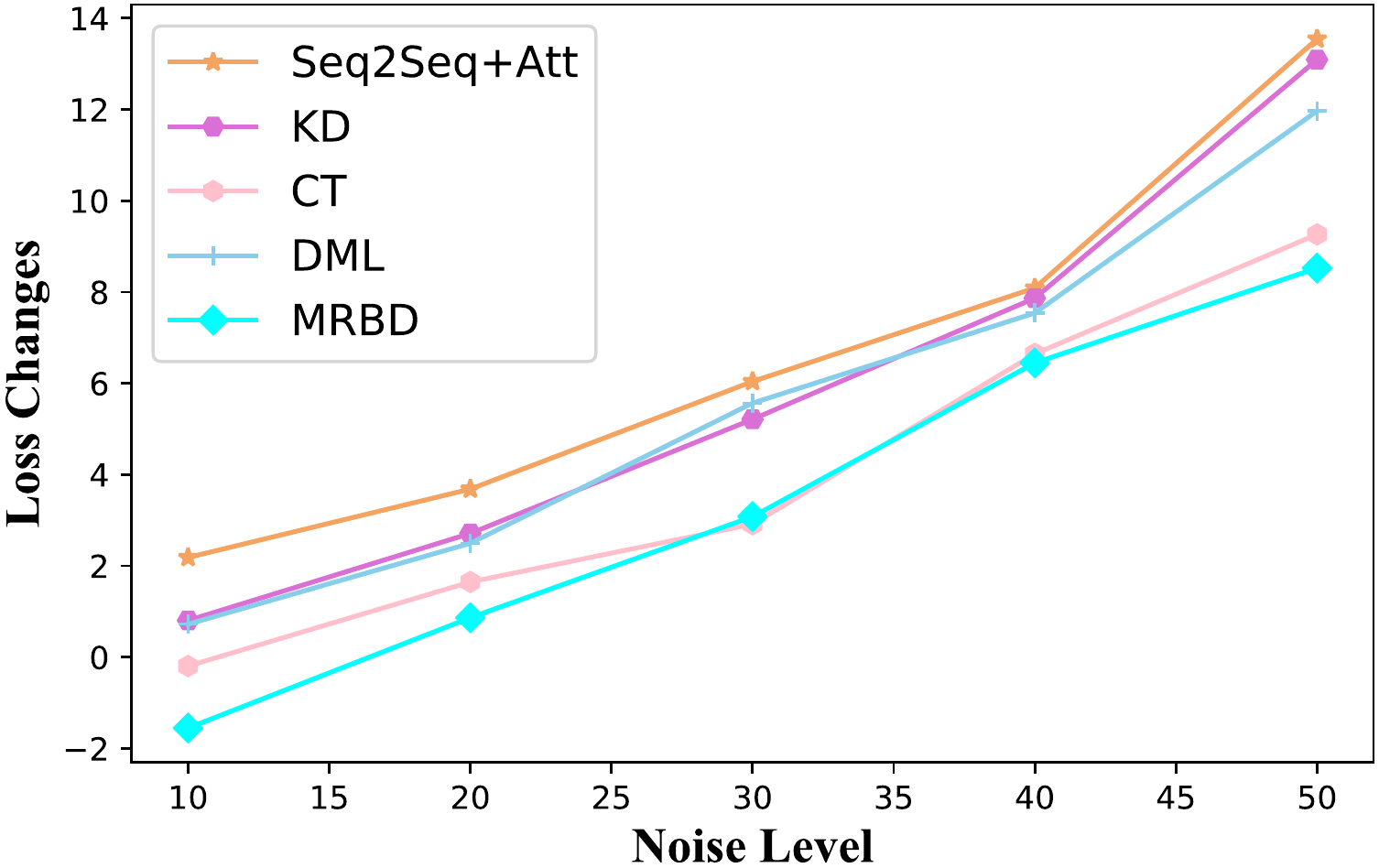}
    \caption{Robustness against noisy data.}
    \label{fig: noisy data robustness}
    % \vspace{-0.5\baselineskip}
\end{figure}
%%%%%%%%%%%%%%%%%%%%%%%%%%%%%%%%%%%%%%%%%%%%%%%%%%%%%%%%%%%%%%%%%%%%%%%%%%%%%%%%%%%%%%%%%%%%%%%
\textbf{Robustness against Noisy Data} 
It is challenging to collect a large-scale and high-quality dialogue dataset.
Moreover, identifying data noise in the raw dialogue dataset is labor-consuming. 
Knowledge distillation is beneficial for the model to resist noisy data because the student not only learns from reference but also considers the prediction from the teacher. 
To evaluate the robustness of models against data noise, we first add noisy data into the training set by replacing the correct responses with randomly selected responses, and then observe the changes of the test loss with respect to the proportion of noisy data.
We report the results in Figure \ref{fig: noisy data robustness}.
Our method and CT achieve better robustness than other baselines, attributed to the independent training sets in CT and the subtask mechanism in MRBD.

%%%%%%%%%%%%%%%%%%%%%%%%%%%%%%%%%%%%%%%%%%%%%%%%%%%%%%%%%%%%%%%%%%%%%%%%%%%%%%%%%%%%%%%%%%%%%%%
\begin{figure}[t]
    \centering 
    \includegraphics[width=0.53\linewidth]{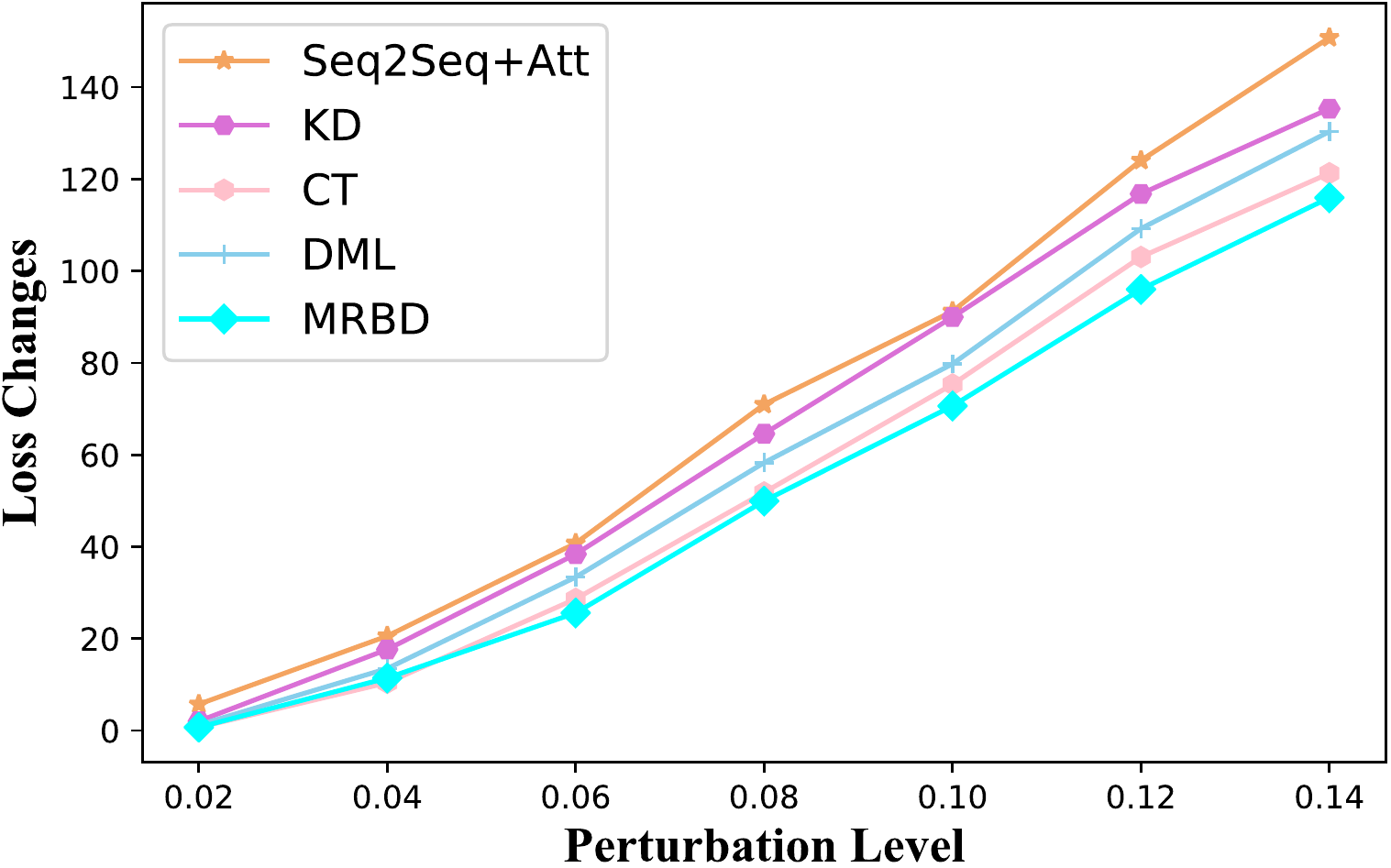}
    \caption{Robustness against parameter perturbation.}
    \label{fig: parameter perturbation robustness}
    % \vspace{-0.5\baselineskip}
\end{figure}
%%%%%%%%%%%%%%%%%%%%%%%%%%%%%%%%%%%%%%%%%%%%%%%%%%%%%%%%%%%%%%%%%%%%%%%%%%%%%%%%%%%%%%%%%%%%%%%
\textbf{Robustness against Parameter Perturbation} 
Previous research \cite{DBLP:conf/iclr/ChaudhariCSLBBC17,DBLP:conf/iclr/KeskarMNST17} has proved that a wider local minimum generally represents better generalization. Specifically, with a wide local minimum, the accuracy of model prediction will not change dramatically under small perturbations in inference. 
To measure the width of local minima reached by baselines and our method, we add independent Gaussian noise with variable standard deviation $\sigma$ to the parameters of the learned models, and then observe the changes of the test loss. 
Figure \ref{fig: parameter perturbation robustness} plots the loss changes with respect to the perturbation level (i.e., the magnitude of $\sigma$). 
We can see that the losses of baselines change more drastically than our method after adding perturbations, which means MRBD finds a wider local minimum than baselines, indicating better generalization.

%%%%%%%%%%%%%%%%%%%%%%%%%%%%%%%%%%%%%%%%%%%%%%%%%%%%%%%%%%%%%%%%%%%%%%%%%%%%
\begin{table}[t]
\centering
\footnotesize
\setlength{\tabcolsep}{4.0pt}

\begin{tabular}{@{}lcccc@{}}
\hline
Metrics & KD & CT & DML & MRBD \\ \hline
Entropy & 1.360 & 5.475 & 4.362 & \bf 5.765 \\
Diversity & 0.580 & 0.747 & 0.706 & \bf 0.798 \\ \hline
\end{tabular}
\caption{Entropy and diversity of predictions generated by baselines and our method.}
\label{Result Analysis: Entropy and Diversity}
% \vspace{-0.5\baselineskip}
\end{table}
%%%%%%%%%%%%%%%%%%%%%%%%%%%%%%%%%%%%%%%%%%%%%%%%%%%%%%%%%%%%%%%%%%%%%%%%%%%%
\subsubsection{Effect of Multi-View Feature Representation} 
Finally, we discuss why multi-view feature representation performs better regularization on the feature learning of students. 

\textbf{Entropy Analysis of Prediction} 
According to previous work \cite{DBLP:conf/iclr/PereyraTCKH17,DBLP:conf/iclr/ChaudhariCSLBBC17}, the softened target with high entropy is beneficial for models to reach a wider local minimum. 
Therefore, we compare our training framework with baselines (except for Seq2Seq+Att) in terms of entropy of predictions of the teacher (or the student peers). 
Note that for MRBD, we choose the first three students to aggregate the predictions. 
The results are reported in Table \ref{Result Analysis: Entropy and Diversity}. Our method obtains the highest entropy, which suggests that multi-view feature representation can enhance the entropy of predictions for better regularization.

\textbf{Diversity Analysis of Prediction} 
We further evaluate the diversity of predictions that reflects the degree of homogenization of students. 
The diversity is calculated by the average Euclidean distance between the predictions of each pair of students. 
As shown in Table \ref{Result Analysis: Entropy and Diversity}, MRBD outperforms all baselines, which indicates that multi-view feature representation effectively alleviates the homogenization of students. 
Besides, the results also demonstrate that the diversity of predictions is related to the entropy of predictions. The predictions with more diversity can avoid the centralization of probability distribution in the aggregated prediction.

%%%%%%%%%%%%%%%%%%%%%%%%%%%%%%%%%%%%%%%%%%%%%%%%%%%%%%%%%%%%%%%%%%%%%%%%%%%%
\begin{table}[t]
\centering
\footnotesize
\setlength{\tabcolsep}{4.0pt}

\begin{tabular}{@{}lcccccc@{}}
\hline
Model & Dist-1 & Dist-2 & Ent-1 & Ent-2 & Dis-1 & Dis-2 \\ \hline
w/ $L_{2}$ & 4.114 & 28.538 & 7.737 & 12.827 & 0.146 & 0.423 \\ \hline
w/ LS & 4.307 & 29.297 & 7.771 & 13.026 & 0.152 & 0.411 \\ \hline
\end{tabular}
\caption{Results of Seq2Seq+Att w/ $L_{2}$ (weight decay), w/ LS (label smoothing).}
\label{Result Analysis: compared with L2 and LS}
% \vspace{-0.5\baselineskip}
\end{table}
%%%%%%%%%%%%%%%%%%%%%%%%%%%%%%%%%%%%%%%%%%%%%%%%%%%%%%%%%%%%%%%%%%%%%%%%%%%%
\subsubsection{Effect of Knowledge for Regularization} 
To validate the effectiveness of knowledge in predictions, we further introduce another kind of regularization approach, including weight decay ($L_{2}$) and label smoothing, to compare with the above four knowledge-based regularization approaches. 
Based on results in Table \ref{experiment result: automatic evaluation} and Table \ref{Result Analysis: compared with L2 and LS}, we can find that the knowledge-based regularization has significant advantages compared with the weight decay-based regularization. 
The label smoothing-based regularization achieves better performance than KD and is slightly weaker than both CT and DML, but MRBD still outperforms it in all metrics. 
Thus knowledge in predictions indeed conducts a more informed guide than the conventional regularization approaches. 

%%%%%%%%%%%%%%%%%%%%%%%%%%%%%%%%%%%%%%%%%%%%%%%%%%%%%%%%%%%%%%%%%%%%%%%%%%%%
\begin{table}[t]
\centering
\footnotesize

\begin{tabular}{@{}p{1\linewidth}@{}}
\hline
\textbf{Dialogue history:} I'd like to apply for a credit card. Can you help me with my application? \\ \hline
\textbf{Seq2Seq+Att:} Then what would you like for a computer? \\
\textbf{KD:} Yes, which size do you recommend? \\
\textbf{CT:} Sir, your account certainly is? You need a credit card. \\
\textbf{DML:} Yes, it's very simple. Would you have an e-mail? \\
\textbf{MRBD:} Certainly, madam. Which one would you like to apply for? \\ \hline \hline
\textbf{Dialogue history:} What subjects are you studying? \\ \hline
\textbf{Seq2Seq+Att:} I am working in Germany and I am studying studying. \\
\textbf{KD:} I am studying in a company. I am studying with you.  \\
\textbf{CT:} I am not interested in English. \\
\textbf{DML:} I have to work abroad and work in English. \\
\textbf{MRBD:} I am studying languages. I am going to work at the university.\\ \hline
\end{tabular}
\caption{Examples generated by baselines and MRBD.}
\label{Result Analysis: Case Study}
% \vspace{-0.5\baselineskip}
\end{table}
%%%%%%%%%%%%%%%%%%%%%%%%%%%%%%%%%%%%%%%%%%%%%%%%%%%%%%%%%%%%%%%%%%%%%%%%%%%%
\subsubsection{Case Study}
Table \ref{Result Analysis: Case Study} shows several examples that consist of dialogue history and responses generated by different models. 
We can see that the responses generated by our method show more relevance with dialogue history than baselines.
Although the knowledge-based baselines also generate some responses that contain words or phrases related to conversation topics, the semantics of these responses are still contradictory to dialogue history.
Besides, these baselines represent low fluency, especially Seq2Seq+Att.
\section{Related Work}

Previous seq2seq-based dialogue models \cite{DBLP:journals/corr/VinyalsL15,DBLP:conf/acl/ShangLL15,DBLP:conf/naacl/SordoniGABJMNGD15,DBLP:conf/aaai/SerbanSBCP16} tend to generate dull and meaningless responses when interacted, although they usually perform well in the training set. 
To tackle this problem, one line of work introduces common knowledge of the real world to constrain the feature learning of models for better generalization. 
\citet{DBLP:conf/naacl/LiGBGD16} first proposed to use mutual information maximization as the training objective instead of only predicting target sequence. \citet{DBLP:conf/emnlp/LiMRJGG16} considered the conversation task as a reinforcement learning problem and used rewards as the training objective. Some work \cite{DBLP:conf/emnlp/LiMSJRJ17,DBLP:conf/nips/ZhangGGGLBD18,DBLP:conf/aaai/FengCLY20} further proposed a variety of manually or automatically defined rewards for more effective and comprehensive constraints. 
Beside, modifying the generation process with task-related inductive biases is also a promising attempt, such as syntactic-based generation \cite{DBLP:conf/acl/DusekJ16,DBLP:conf/icml/WelleckBDC19}, hierarchical generation \cite{DBLP:conf/aaai/SerbanKTTZBC17,DBLP:conf/naacl/SuLYC18}, and latent variable based generation \cite{DBLP:conf/aaai/SerbanSLCPCB17,DBLP:conf/acl/ZhaoZE17,DBLP:conf/iclr/GuCHK19,DBLP:conf/acl/ShenFZ19}. 
Several research even incorporated more specific knowledge into the dialogue task, such as topics \cite{DBLP:conf/aaai/XingWWLHZM17}, personas \cite{DBLP:conf/ijcai/QianHZXZ18,DBLP:conf/acl/KielaWZDUS18}, emotions \cite{DBLP:conf/aaai/ZhouHZZL18}, implicit scenarios \cite{DBLP:conf/emnlp/FengRCSLS20}, and structure knowledge \cite{DBLP:conf/aaai/GhazvininejadBC18,DBLP:conf/aaai/YoungCCZBH18,DBLP:conf/sigir/ZhanZCSLDY20}.

Our work belongs to another line of work that aims to capture common knowledge from the training data by knowledge distillation (KD) \cite{DBLP:journals/corr/HintonVD15}. 
\citet{DBLP:conf/sigir/TahamiGS20} introduced KD into the retrieval-based dialogue model where knowledge from a better performing teacher is used to regularize a lower-performance but much faster student. 
\cite{DBLP:conf/acl/FengTWFZY19} proposed a Co-Teaching retrieval-based dialogue model where two students are optimized on independent training sets, and imitate and guide each other with their own knowledge from the assigned training sets. 
However, both KD and Co-Teaching limit the teacher knowledge to single-view feature representation. As the training data is shuffled once in each epoch, each student in Co-Teaching can still access the data allocated to another student, which means two students may learn similar feature representations. Our work is more related to deep mutual learning (DML) \cite{DBLP:conf/cvpr/ZhangXHL18} where a group of students with different initializations even architectures are trained on the same dataset and try to learn multi-view knowledge. Each student aggregates the teacher knowledge from all other students equally. Unfortunately, students tend to learn similar feature representations due to optimized on the same dataset, and will further homogenize based on similar teacher knowledge. These problems hinder students from learning diverse views. Meanwhile, the computation cost will increase dramatically as the number of views grows. More importantly, all of the above methods still conduct unidirectional knowledge distillation, which is not beneficial for continuous performance improvement.

The existing NLG tasks using KD mainly contain neural machine translation and text generation \cite{DBLP:conf/emnlp/KimR16,DBLP:conf/acl-deeplo/TangLL19,DBLP:conf/naacl/WeiHWDC19,DBLP:conf/acl/ChenGCLL20}. 
To the best of our knowledge, our method, including multi-view feature representation and bidirectional distillation, is the first work that applies knowledge distillation to generative dialogue systems.

\section{Conclusion}
In this work, we propose a novel training framework, multi-view feature representation with bidirectional distillation (MRBD), to guide the dialogue model towards better generalization. The students in MRBD not only learn from the assigned subtasks but also imitate diversified multi-view knowledge from the randomly selected student peers trained on different unseen subtasks. Besides, we further construct bidirectional distillation that allows the student peers to exchange knowledge simultaneously and find common parts together. Therefore, the proposed method can automatically capture common knowledge by maintaining a balance between the diversity and consistency of feature representation. The experimental results and analysis validate the superiority of the knowledge-based regularization and demonstrate the effectiveness of multi-view feature representation and bidirectional distillation.

\section{Acknowledgements}
This research is supported by Beijing Natural Science Foundation (No. L181010 and 4172054), National Key R\&D Program of China (No. 2016YFB0801100), National Basic Research Program of China (No. 2013CB329605), and Beijing Academy of Artificial Intelligence (BAAI). Xu Sun and Kan Li are the corresponding authors.

\bibliography{aaai}
\bibstyle{aaai21}

\end{document}